\documentclass[10pt]{article}
\usepackage[preprint]{neurips_2026}

\makeatother
\usepackage{booktabs}
\usepackage{multirow}

\usepackage{amsmath,amssymb}
\usepackage{graphicx}
\usepackage{xcolor}
\usepackage{hyperref}
\usepackage{wrapfig}
\usepackage{soul}

\usepackage{amsthm}

\newcommand{\real}{\ensuremath{\mathbb{R}}}

\newcommand{\diff}{\mbox{Diff}^+(D)}

\usepackage{titlesec}
\titlespacing*{\section}{0pt}{1.0ex}{0.6ex}
\titlespacing*{\subsection}{0pt}{0.8ex}{0.4ex}

\title{Bayesian \textit{In Vivo} Tracking of Synapses using Joint Poisson Deconvolution and Diffeomorphic Registration}

\author{
Shashwat Kumar$^{1,*}$ \quad
Dominic M. Padova$^{1,*}$ \quad
Binish Narang$^{1}$ \quad
Gabrielle I. Coste$^{2}$ \quad\\
\textbf{Austin R. Graves}$^{2,3}$ \quad
\textbf{Richard L. Huganir}$^{2,3}$ \quad
\textbf{Adam S. Charles}$^{1,3,4}$ \quad
\textbf{Michael I. Miller}$^{1,3}$ \quad\\ \textbf{Anuj Srivastava}$^{4,5}$\\
$^{1}$Department of Biomedical Engineering, Johns Hopkins University \\
$^{2}$Department of Neuroscience, Johns Hopkins University \\
$^{3}$Kavli Neuroscience Discovery Institute, Johns Hopkins University \\
$^{4}$Data Science and AI Institute, Johns Hopkins University \\
$^{5}$Department of Applied Mathematics and Statistics, Johns Hopkins University\\
$^{*}$Equal contribution
}
\begin{document}
\maketitle
\footnotetext[1]{These authors contributed equally to this work.}
\begin{abstract}
Synapses are densely packed submicron structures that dynamically reorganize during learning and memory formation. Longitudinal \textit{in vivo} imaging of fluorescently tagged synaptic receptors offers a promising opportunity to study large-scale synaptic dynamics and how these processes are disrupted in neurological disease. However, in vivo imaging with 2-photon microscopy uses low laser power and therefore suffers from low signal-to-noise ratio (SNR) and high shot noise, nonlinear tissue motion between days, nonstationary fluctuations in synaptic fluorescence, and significant blur induced by the microscope point spread function (PSF). Together, these factors make it challenging to detect and track synapses, especially in regions with high synaptic density. This paper presents a novel template-based framework for modeling synapses as varying luminance point sources that move under a nonlinear tissue deformation. Taking a unified Bayesian approach, we apply this model to microscopy data by deriving a posterior that incorporates a diffeomorphic mapping for domain warping, a Gaussian point spread function for the imaging process, and a Poisson observation model for raw photon counts. The Bayesian solution simultaneously: (1) Constructs a probabilistic template of synapse locations, (2) denoises and deconvolves the image data, (3) infers fluorescence intensities, (4) performs diffeomorphic image registration to correct for tissue motion, and (5) provides confidence regions for these parameter estimates. We demonstrate the framework on both a 2D+t simulated dataset and a 3D+t longitudinal \textit{in vivo} microscopy dataset of fluorescent synapses imaged in a mouse over two weeks.
\end{abstract}
\section{Introduction}
\begin{figure}[!ht]
    \centering
    \includegraphics[width=0.95\linewidth]{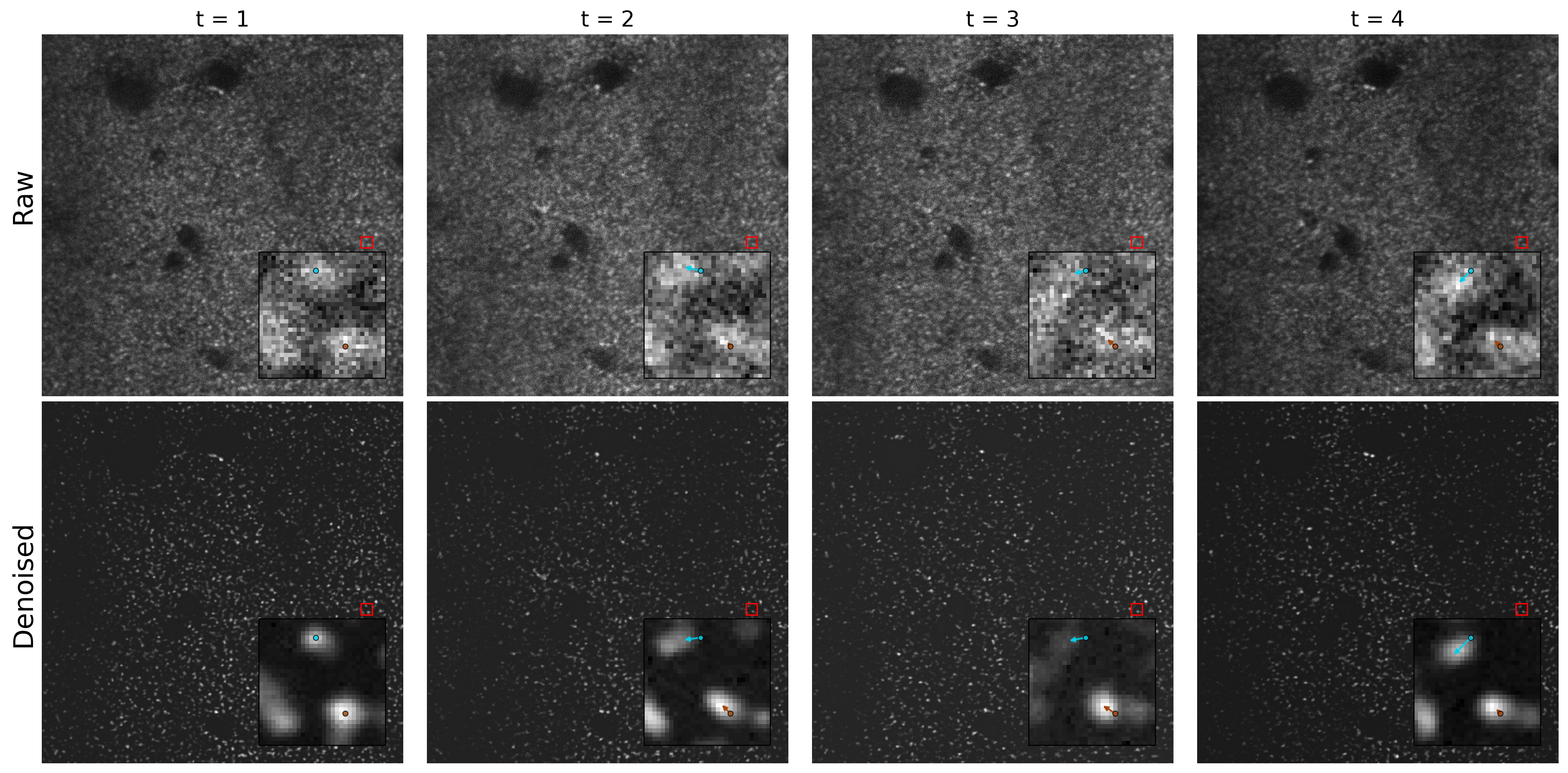}

    \caption{Longitudinal \textit{in vivo} synapse imaging is noisy and exhibits nonlinear motion.
(a) Raw XY slices across days show low signal-to-noise ratio and substantial shot noise.
(b) Output of a denoising autoencoder (XTC). Insets visualize synaptic motion by comparing their positions with first day.}
    \label{fig:introduction}
\end{figure}

Synapses are the fundamental unit of neural connectivity and communication that enable animals to learn, adapt, and form memories~\citep{malinow2002ampa,morris1989synaptic}. Recent advances in transgenic mouse lines and in vivo imaging technologies have made it possible to longitudinally image hundreds of thousands of synapses in vivo~\citep{graves2021visualizing} (Fig.~\ref{fig:introduction}). Detecting and tracking synaptic changes at this scale could reveal the mechanisms underlying learning and adaptive computation in living animals, and potentially lead to new learning rules for artificial systems. However, large-scale longitudinal synapse tracking remains elusive. Synaptic fluorescence is non stationary across imaging sessions due to biological plasticity and imaging variability. Synapses also move with dendritic branches and surrounding tissue, leading to nonlinear tissue deformation between observations. Additionally, longitudinal deep tissue imaging requires lower laser power to avoid photobleaching of tissue, leading to poorer SNR and higher shot noise (Fig.~\ref{fig:introduction}, row 1).

\begin{figure*}[!ht]\includegraphics[
  width=\linewidth]{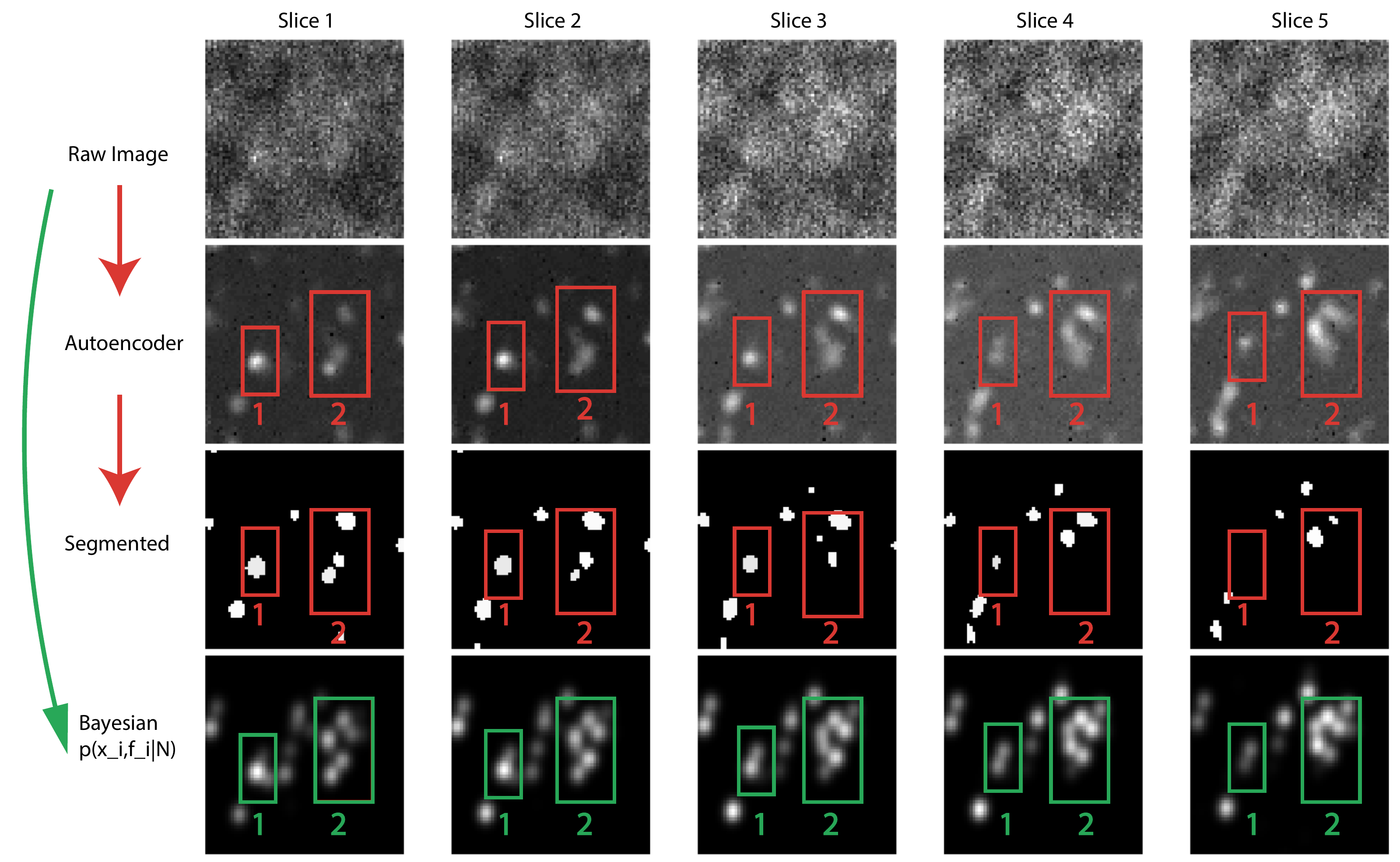}
    \caption{Failure modes of the denoising-segmentation-tracking pipeline. Multiple XY slices at different depths ($z$-planes) from a small 3D image volume are shown. The denoising autoencoder frequently blurs adjacent synapses. Constellation 1 contains two synapses (Autoencoder, Slice 4) that are merged into a single detection by the segmentation stage. Constellation 2 contains a cluster of six synapses that become blurred by the autoencoder in Slices 4 and 5, again leading to merge errors. In contrast, the Bayesian model is able to deconvolve these touching synapses.}
    \label{fig:merge_errors}
\end{figure*}

While past works have shown promising results by combining denoising~\citep{xu2023cross}, segmentation~\citep{chen2025automatic}, and tracking~\citep{kumar2025uncertainty}, these tracking-by-detection approaches can fail in crowded or low-SNR regimes and do not explicitly model nonlinear tissue deformation (Fig.~\ref{fig:merge_errors}). In particular, such pipelines typically rely on denoising autoencoders followed by CNN-based classification and thresholding to produce discrete detections. As illustrated in Figure~\ref{fig:merge_errors}, denoising can blur nearby synapses in crowded regions, causing downstream segmentation networks to merge closely spaced synapses into a single structure. Errors introduced during the denoising $\rightarrow$ detection $\rightarrow$ tracking pipeline can then propagate across stages and become difficult to correct retrospectively. Furthermore, object detectors make local decisions over limited spatiotemporal neighborhoods, which can bias detections toward brighter synapses, skew fluorescence estimates, and discard uncertainty information.

To address these challenges, we draw inspiration from a similar problem in astrophysics: catalogue building. Here, given images of the sky, astronomers either hand engineer or use Markov Chain Monte Carlo (MCMC) to build a star catalogue directly from images~\citep{brewer2013probabilistic}. Unlike astronomical scenes, however, synapses move across imaging sessions due to nonlinear tissue deformation. To model this motion, we use tools from computational anatomy~\citep{grenander1998computational} and represent synapses as dynamic catalogue of point sources moving under a smooth topology-preserving tissue deformation. By allowing the same catalogue of point sources to evolve under the deformation field, the model consistently explains observed image peaks across time while maintaining coherent synapse identities. We jointly infer synapse locations and trajectories, synaptic fluorescence, and diffeomorphic motion fields, directly from noisy images, rather than follow the status quo denoise-segment-track pipeline which discards uncertainty, compounds errors, and can fail under crowding and deformation.

\textbf{Contributions:}
\begin{itemize}

\item \textbf{Tracking-by-synthesis with a diffeomorphic template.} 
We fit a latent catalogue of point sources "wiggling" under the effect of diffeomorphic transformations to synthesize virtual synapses which can provide temporally coherent explanations of longitudinal microscopy images.

\item \textbf{Physics-based generative model of image formation.} 
Our model bakes in the imaging physics with an explicit parameterization of the Point Spread Function while modeling both foreground signal and background shot noised based on Poisson statistics.

\item \textbf{Joint Bayesian inference with uncertainty.} 
By comparing the synthesized volumes with real data, our model jointly performs registration, builds the catalogue, denoises the data while providing calibrated uncertainty estimates for all parameters.

\end{itemize}

\section{Related Work}
Longitudinal \textit{in vivo} synapse tracking has historically relied on sparse labeling and manual analysis because dense 3D synapse imaging is noisy, crowded, and undergoes nonlinear deformation across imaging sessions~\citep{roth2020cortical}. Large-scale longitudinal reconstruction of neurons has recently become possible~\citep{winnubst2019reconstruction}, suggesting that analogous large-scale synapse tracking may also be achievable. Recent advances in transgenic mouse models and multiphoton microscopy have enabled longitudinal \textit{in vivo} imaging of hundreds of thousands of synapses, opening new opportunities for studying synaptic plasticity at scale~\citep{graves2021visualizing}. Existing computational pipelines typically decompose the problem into separate denoising~\citep{xu2023cross}, segmentation~\citep{chen2025automatic}, registration, and tracking~\citep{kumar2025uncertainty} stages.

As an alternative to the detection paradigm, sparse deconvolution approaches model fluorescence images as convolutions of latent point sources with a microscope point spread function (PSF), often using continuous-domain regularization over Radon measures for super-resolution recovery~\citep{denoyelle2017support} or Bayesian sparse inference formulations~\citep{park2013variational}. Related work in microscopy has also explored Bayesian and variational image restoration, multiview deconvolution, and PSF estimation for fluorescence and multiphoton imaging~\citep{preibisch2013efficient,ajdenbaum2023variational}. Probabilistic catalogue construction in astrophysics similarly infers latent point sources directly from noisy telescope images using Bayesian inference and MCMC methods~\citep{brewer2013probabilistic,regier2019approximate}. These approaches jointly model image formation, overlapping point spread functions, and uncertainty in source properties. Unlike astronomical scenes, however, longitudinal synapse imaging additionally requires modeling nonlinear tissue deformation and temporal evolution across imaging sessions.

In parallel, diffeomorphic registration methods have become central to computational anatomy and nonlinear image alignment~\citep{beg2005computing,joshi2000landmark,glaunes2004diffeomorphic,trouve2005metamorphoses,younes2010shapes}. Unlike prior approaches that operate on post-detection point estimates or perform dense image registration independently, our framework jointly models latent fluorescence sources, optical image formation, uncertainty quantification, and nonlinear longitudinal deformations within a unified probabilistic inverse model.

\section{Bayesian Formulation and MCMC Inference}
In this section, we represent images as noisy outcomes of a Poisson point process observed through diffeomorphic shifts and point spread functions of the microscope. We impose statistical models on unknown variables and derive a joint posterior. Finally, we develop an MCMC process to generate inferences and estimate synapse localizations. 

\subsection{Mathematical Representation of Synapse Observations}
We start by deriving a mathematical model for the synapse generation process and its observations. 

{\bf Synapse Locations and Fluorescence}: Let $\{x_i \in D \subset \real^3\}$ denote synapse locations in the observed domain $D$. $ D \subset \mathbb{R}^3$ is a volume representing the imaged space; for convenience, we will assume $D = [0,1]^3$. Let the synapses on day $t$ be represented by a singular density $\mu_t(x) = \sum_{i=1}^I f_{i,t}\delta(x-x_i)$. Here $f_{i,t} \in \mathbb{R}_{\geq 0}$ denotes the fluorescence or the intensity of $i^{th}$ synapse on day $t$. 

{\bf Diffeomorphic Domain Shifts}: We model the nonlinear shifts on day $t$ by a diffeomorphism of the domain: $\varphi_t: D \to D$. Let $\diff$ denote the group of all positive diffeomorphisms of $D$ and let $\varphi_I$ denote the identity element of the group. $T_{\varphi_I}(\diff)$ denotes the Lie algebra of $\diff$. It can be shown that $T_{\varphi_I}(\diff)$ is the set of all boundary-preserving smooth vector fields on $D$, {\it i.e.}, these vector fields are tangential on $\partial D$, the boundary set of $D$~\citep{ebin1970groups, younes2010shapes}. The diffeomorphism $\varphi_t$ acts on the synapse measure by the left action:
\begin{align}
(\varphi_t \circ \mu_t)(x) := \sum_{i=1}^I f_{i,t} \delta(x-\varphi_t(x_i))\ .
\end{align}
If we assume that the diffeomorphisms $\varphi_t$ are small, {\it i.e.}, close to $\varphi_I$ in the group $\diff$, they can be approximated by displacement fields $u_t \in T_{\varphi_I}(\diff)$ satisfying Neumann boundary conditions. In other words, the diffeomorphism $\varphi_t: D \to D$ is given by $\varphi_t(x) \approx x + u_t(x)$. In practice, the displacement field $u_t$ can be compactly represented in a reproducing kernel Hilbert space of vector fields according to $u_t(x) = \sum_{i=1}^I K(x,x_{i}) p_{i,t}$. Here, $\{p_{i,t} \in \real^3\}$ denote momenta (cotangent coefficient vectors) at synapse locations $\{ x_i \in D\}$. This representation reduces the original infinite-dimensional estimation of $\varphi_t \in \diff$ to a finite-dimensional parameterization of the displacement field.
\paragraph{Microscope PSF} 
We denote the point spread function of the microscope camera by $\psi: \mathbb{R}^3 \to \mathbb{R}$, and model it as a Gaussian density: 
$$
\psi(x) = (2\pi)^{-\frac{3}{2}}|\Sigma_{\text{PSF}}|^{-\frac{1}{2}}e^{-\frac{1}{2}x^T \Sigma_{\text{PSF}}^{-1} x}.
$$
In this paper we use a fixed $\Sigma_{\text{PSF}}$ estimated from data segmented using a previous pipeline~\citep{chen2025automatic}. Under these assumptions, the mean or expected image of a synapse field is the result of the  convolution:
\begin{equation}
\lambda_t(x) = (\psi * (\varphi_t \circ \mu_t))(x) =  \sum_{i=1}^I f_{i,t} \psi(x-\varphi_t(x_i))\ . \label{eq:obs-model}
\end{equation}
An actual image is a pixel-wise readout of this mean image, with each pixel denoting a random count derived from the corresponding mean pixel. 

\paragraph{Model parameters}
The collection of all unknown variables is given by
\begin{equation}
\theta = \Big\{ \{x_i\}_{i=1}^I,\ \{f_{i,t}\}_{i=1,t=1}^{I,T},\ \{p_{i,t}\}_{i=1,t=1}^{I,T},\ \lambda_0 \Big\}
\in \Theta \triangleq  D^I \times \mathbb{R}_{\geq 0}^{IT} \times \mathbb{R}^{3IT} \times \mathbb{R}_{\geq 0}.
\end{equation}
Here, $m \triangleq \dim(\theta) = 3I + IT + 3IT + 1$. 
Next, we impose prior models on $\theta$ and reach a posterior distribution on these variables using the Bayes' rule. 

\subsection{Posterior Modeling}
We make the following prior assumptions about the model parameters. 

\begin{itemize}
\item {\bf Synapse locations} are assumed to be independently and uniformly distributed over the domain:
$x_i  \overset{\text{i.i.d.}}{\sim} \mathrm{Uniform}(D)$.

\item {\bf Fluorescence intensities} are modeled as independent exponential random variables:
$f_{i,t} \overset{\text{i.i.d.}}{\sim} \mathrm{Exponential}(\kappa)$.
This choice encourages sparsity in fluorescence, allowing synapses to effectively appear or disappear across time. 

\item To model {\bf nonlinear shifts of synapses}, we parameterize diffeomorphic deformations using sparse momenta variables $p_{i,t} \in \mathbb{R}^3$, which are assumed to be independent Gaussian:
$p_{i,t}  \overset{\text{i.i.d.}}{\sim} \mathcal{N}(0, \Sigma_{\text{motion}})$.
Through the kernel representation of the displacement field, this induces smooth spatial deformations. While more structured priors (e.g., spatiotemporal coupling) could be considered, we adopt an {\it i.i.d.}\ model for simplicity.

\item {\bf Background intensity.}
We place a half-normal prior on the background fluorescence $\lambda_0 \sim \mathrm{HalfNormal}(\nu)$ with log-density $\log p(\lambda_0) = -\lambda_0^2/(2\nu^2) + \text{const}$.

\item  Let $\Omega \subset D$ denote the discrete voxel grid at which photon counts are observed. The {\bf observed counts} $\{N_t(x), x \in \Omega\}_{t=1}^T$ are modeled as Poisson random variables resulting in the log-likelihood:
\begin{align}
\log L(\{N_t(x)\} \mid \{\lambda_t(x)\}, \lambda_0)
&= \sum_{t=1}^T\sum_{x \in \Omega}\log L(N_t(x) \mid \lambda_t(x), \lambda_0) \nonumber \\ 
&\propto \sum_{t=1}^T\sum_{x \in \Omega}
-(\lambda_t(x) + \lambda_0) + N_t(x) \log(\lambda_t(x) + \lambda_0) \ .
\label{eqn:log-likelihood}
\end{align}
where $\lambda_t(x)$ is as given by Eqn.~\ref{eq:obs-model}.
\end{itemize}

Combining the prior terms with the likelihood function we obtain the log posterior: 
\begin{align}
\log\pi(\theta| \{N_t\}_{t=1}^T) &\propto -U(\theta) =   
IT\log(\kappa) -\sum_t \sum_i \kappa f_{i,t} -
\frac{1}{2} \sum_t \sum_i p_{i,t}^T \Sigma_{\text{motion}}^{-1} p_{i,t} \nonumber \\
&-\frac{\lambda_0^2}{2\nu^2} + \log L(\{N_t(x)\} \mid \{\lambda_t(x)\}, \lambda_0)\ ,
\end{align}
where $\log L$ is given above in Eqn.~\ref{eqn:log-likelihood}.

\subsection{Posterior Inference: Hamiltonian Monte Carlo}
Since this posterior is complex due to the nonlinear deformation and convolution operations, we sample it using Hamiltonian Monte Carlo method. We construct a Hamiltonian function:
\begin{equation}
    H(\theta,v) = U(\theta) + \frac{1}{2} v^T M^{-1} v\ .\nonumber
\end{equation}
Here $v$ is a vector of dimension $m = \dim(\theta)$ and $M$ is a $m \times m$ symmetric positive-definite matrix. 
Let $\theta \in \Theta$ denote the current point in the search space. 
Then, under HMC, one uses Hamilton's equations of motion to generated proposals for $\theta$ and $v$: 
$$
\dot{\theta} = \frac{\partial H}{\partial v} \in \ \real^m,\ \ \dot{v} = -\frac{\partial H}{\partial \theta} \in \ \real^m\ .\nonumber
$$
The gradient of $U$ {\it w.r.t} $\theta$ is given by: 
$    \nabla_{\theta} U = \begin{bmatrix}
        \nabla_{f} U,
        \nabla_{x} U,
        \nabla_{p} U,
        \nabla_{\lambda_0} U
    \end{bmatrix} \in \real^m$. 
Let $r_t(x, \lambda_0) = (\frac{N_t(x)}{\lambda_t(x) + \lambda_0} - 1) \in \real$ denote the Poisson residual. The expression for the desired gradients can be expressed in terms of their partial derivatives:


\begin{align}
\frac{\partial U}{\partial f_{i,t}}
&= \sum_{x \in \Omega}
-r_t(x, \lambda_0)\, \psi\big(x - \varphi_t(x_i)\big)
+
\kappa \ \ \ \in \real, \\
\frac{\partial U}{\partial x_i}
&=
\sum_{t=1}^T\sum_{x \in \Omega} r_t(x, \lambda_0) f_{i,t} \nabla \psi(x-\varphi_{t}(x_i))^T (I_3 + \sum_{i'=1}^I \nabla_1 k(x_i,x_{i'}) p_{i',t}^T) \ \ \in \real^3, \\
\frac{\partial U}{\partial p_{j,t}}
&= \sum_{x \in \Omega} r_{t}(x, \lambda_0) \sum_{i=1}^I f_{i,t} \nabla \psi(x-\varphi_{t}(x_i))^T k(x_i,x_j) + p_{j,t}^T\Sigma_{\text{motion}}^{-1} \ \ \in \real^3, 
\\
    \frac{\partial U}{\partial \lambda_0} &= -\sum_{t=1}^T \sum_{x \in \Omega} r_t(x, \lambda_0) + \frac{\lambda_0}{\nu^2}
    \label{eq:gradient_lambda_0} \ \ \in \real.
\end{align}

\paragraph{Leapfrog updates.}
Given step size $\epsilon$ and mass matrix $M$, initialize $(\theta^{(0)}, v^{(0)})$. For $l=0,..,L-1$:
\begin{align}
v^{(l+\frac{1}{2})} 
&= v^{(l)} - \frac{\epsilon}{2}\,\nabla_\theta U(\theta^{(l)}), \nonumber\\
\theta^{(l+1)} 
&= \theta^{(l)} + \epsilon\, M^{-1}v^{(l+\frac{1}{2})}, \nonumber\\
v^{(l+1)} 
&= v^{(l+\frac{1}{2})} - \frac{\epsilon}{2}\,\nabla_\theta U(\theta^{(l+1)}).\nonumber
\end{align}
The proposal $(\theta^*, v^*) = (\theta^{(L)}, v^{(L)})$ is accepted with probability
\begin{equation}
\alpha = \min\left(1,\;
\exp\big(-H(\theta^*,v^*) + H(\theta^{(0)},v^{(0)})\big)
\right).\nonumber
\end{equation}

Each HMC iteration consists of:
(i) sampling $v \sim \mathcal{N}(0, M)$,
(ii) performing $L$ leapfrog steps with step size $\epsilon$, and
(iii) accepting or rejecting the proposal via the Metropolis criterion above. In practice, we implement inference using the No-U-Turn Sampler (NUTS) in PyMC~\citep{abril2023pymc}, which adaptively tunes the step size and mass matrix during burn-in to target an acceptance probability of $0.9$. We use 1000 warm-up iterations followed by 1000 posterior sampling iterations.

\section{Experimental Results}
In this section, we present results on sampling-based estimation of model parameter $\theta$ using HMC. We evaluate the proposed method on both simulated and real in vivo synapse imaging data. The simulated dataset is generated from the proposed generative model, providing full ground truth for synapse locations, fluorescence, and deformation fields. The real dataset consists of two-photon images of fluorescently labeled synapses from the SEP-GluA2 transgenic mouse line, which enables brain-wide visualization of GluA2-containing excitatory synapses~~\citep{graves2021visualizing,xu2023cross}. In this model, endogenous AMPAR GluA2 subunits are tagged with the green fluorophore Super Ecliptic pHluorin (SEP), producing dense fields of fluorescent puncta under 910 nm excitation through a cranial window. Imaging was performed over a $100 \times 100 \times 60\,\mu$m volume across eight sessions spanning two weeks, resulting in a challenging setting with low SNR, substantial shot noise, and nonlinear tissue motion.

\subsection{Simulated Data}

\begin{table}[t]
\centering
\caption{Performance on simulated data across varying numbers of synapses ($I$) and time points ($T$) (median [25th, 75th percentiles]).
Increasing $T$ improves template localization and reduces ambiguity through temporal coupling.
In contrast, increasing $I$ degrades performance due to overlap in the point spread function.}
\label{tab:results}
\small
\resizebox{0.98\linewidth}{!}{
\begin{tabular}{c c c c c c c c}
\hline
I & T & Corr $\uparrow$ & Fluorescence (abs) & Template error & Deformation error & Momenta error & Log-likelihood\\
\hline
2 & 4 & 0.98 [0.92, 0.99] & 18.8 [13.8, 30.7] & 2.44 [1.68, 3.36] & 5.60 [1.59, 13.36] & 2.76 [2.17, 3.34] & -11.8 [-12.4, -11.1] \\
2 & 6 & 0.99 [0.84, 0.99] & 17.6 [14.0, 57.1] & 2.01 [1.30, 3.16] & 7.37 [1.39, 13.75] & 2.60 [2.00, 2.99] & -11.7 [-12.6, -11.1] \\
2 & 8 & 0.99 [0.87, 0.99] & 19.1 [14.9, 46.4] & 1.55 [1.16, 2.50] & 7.86 [1.37, 14.68] & 2.34 [1.77, 2.83] & -11.7 [-12.4, -11.0] \\
2 & 10 & 0.99 [0.88, 0.99] & 18.6 [14.4, 52.3] & 1.61 [1.05, 2.40] & 4.42 [1.32, 14.48] & 2.30 [1.81, 2.82] & -11.7 [-12.6, -11.0] \\
4 & 4 & 0.86 [0.62, 0.97] & 49.9 [27.5, 80.1] & 2.90 [2.15, 3.83] & 11.69 [9.05, 16.24] & 2.86 [2.65, 3.35] & -9.1 [-10.1, -8.3] \\
4 & 6 & 0.85 [0.64, 0.96] & 49.8 [30.9, 81.4] & 2.37 [1.68, 3.71] & 12.68 [9.28, 15.40] & 2.81 [2.42, 3.22] & -9.2 [-10.1, -8.3] \\
4 & 8 & 0.82 [0.63, 0.93] & 53.9 [35.4, 88.2] & 2.30 [1.45, 3.68] & 12.07 [8.75, 16.10] & 2.88 [2.42, 3.49] & -8.9 [-9.8, -8.2] \\
4 & 10 & 0.87 [0.73, 0.97] & 45.2 [28.5, 72.2] & 1.91 [1.29, 2.86] & 12.30 [9.11, 16.35] & 2.70 [2.37, 3.14] & -8.9 [-9.7, -8.3] \\
8 & 4 & 0.59 [0.46, 0.69] & 98.1 [81.3, 116.9] & 4.29 [3.71, 5.34] & 14.22 [11.21, 15.77] & 3.15 [2.97, 3.52] & -7.2 [-8.5, -6.6] \\
8 & 6 & 0.63 [0.57, 0.76] & 89.9 [71.6, 107.6] & 3.49 [2.77, 4.39] & 14.31 [12.33, 16.63] & 3.15 [2.93, 3.39] & -6.8 [-7.8, -6.2] \\
8 & 8 & 0.67 [0.55, 0.77] & 85.2 [69.8, 103.1] & 3.16 [2.27, 4.13] & 14.48 [11.93, 16.40] & 3.12 [2.90, 3.38] & -6.8 [-7.5, -6.3] \\
8 & 10 & 0.67 [0.55, 0.76] & 83.4 [72.7, 109.5] & 3.06 [2.11, 4.80] & 14.41 [12.46, 16.45] & 3.23 [2.96, 3.41] & -6.7 [-7.0, -6.3] \\
\hline
\end{tabular}}
\label{table:simulated}
\end{table}

\begin{figure*}[!ht]\includegraphics[
  width=\linewidth]{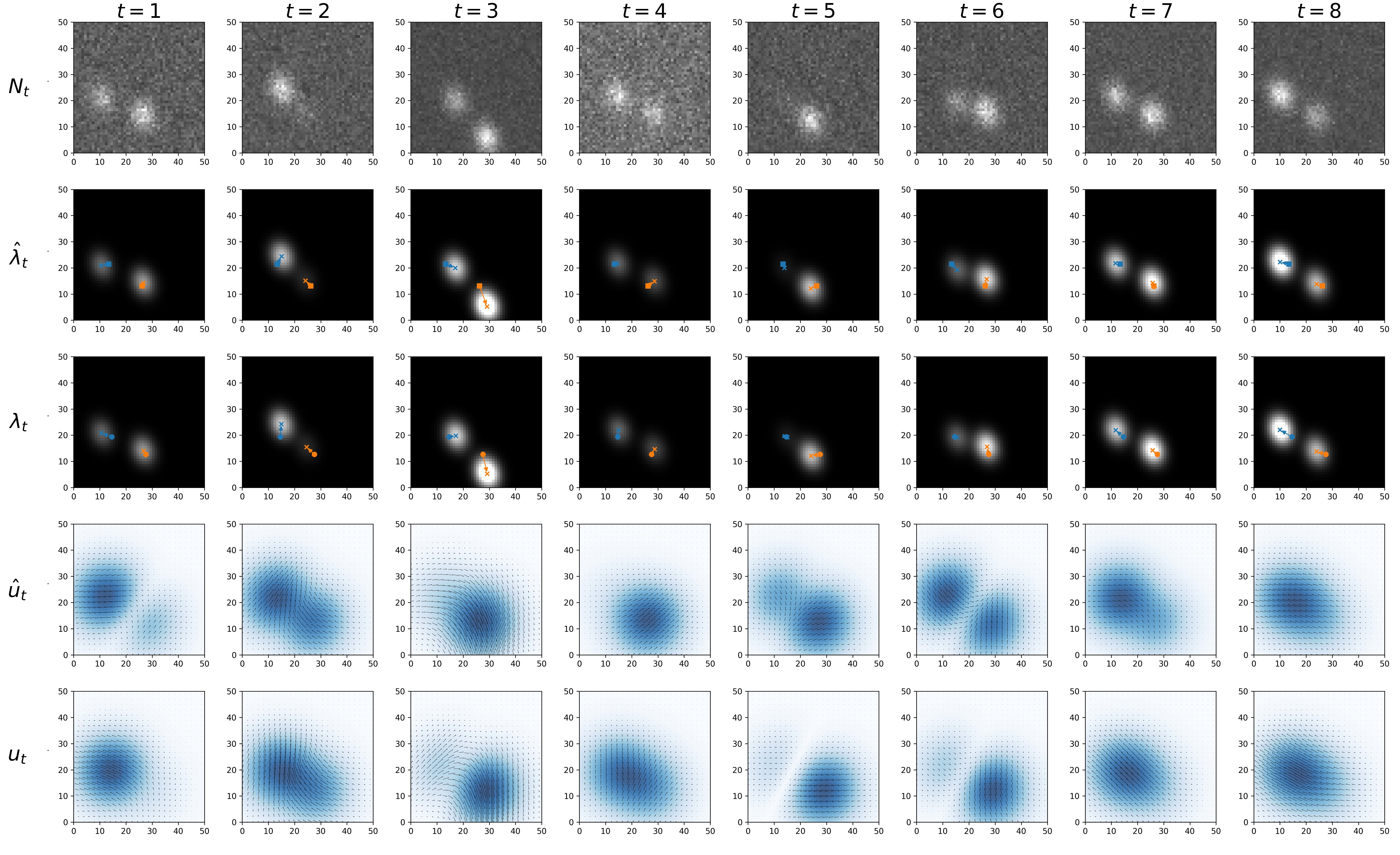}
    \caption{Accurate recovery of signal and motion in simulated data.
Observed photon counts $N_t$ are generated from blurred point sources under Poisson noise.
The inferred intensity $\hat{\lambda}_t$ closely matches the ground-truth $\lambda_t$, with accurate recovery of synapse locations.
Estimated displacement fields $u_t(x)$ agree with the ground-truth motion, demonstrating that the model jointly recovers fluorescence and nonlinear deformation.}
    \label{fig:simulated_data}
\end{figure*}

We synthesize 2D images from the prior using
$\kappa=200$,
$\Sigma_{\text{PSF}}=\begin{bmatrix}
    10 & -2\\
    -2 & 15
\end{bmatrix}$,
and
$\Sigma_{\text{motion}} = \begin{bmatrix}
    3^2 & 0\\
    0 & 3^2
\end{bmatrix}$.
This setup generates sparse point sources undergoing smooth nonlinear motion, observed through a blurred imaging system with Poisson noise, closely matching the assumptions of our generative model.

Figure~\ref{fig:simulated_data}, Row 1 shows simulated photon counts generated from the model. Rows 2 and 3 show the estimated intensity function $\hat{\lambda}_t(x)$ and the ground-truth intensity $\lambda_t(x)$. True template particle locations are shown with circles and estimated particle locations with squares. The close agreement between $\hat{\lambda}_t(x)$ and $\lambda_t(x)$, together with accurate recovery of particle locations, indicates that the model is able to effectively invert the imaging process despite substantial blur and shot noise.

Rows 4 and 5 show the estimated and ground-truth displacement fields, $\hat{u}_t(x)$ and $u_t(x)$ respectively. The strong correspondence between these fields demonstrates that the model can jointly recover nonlinear motion and signal, rather than relying on a separate registration stage. Figure~\ref{fig:simulate_posterior} in the appendix shows posterior distributions obtained from MCMC. The concentrated posterior mass around the ground-truth parameters suggests stable inference and good identifiability in the sparse regime.

Table~\ref{table:simulated} provides quantitative evaluation of these results. All errors are computed after optimal matching between true and estimated template points. We define the template error as $\frac{1}{I} \sum_i \|x_i - \hat{x}_{\pi(i)}\|$, the fluorescence error as $\frac{1}{IT} \sum_{i,t} |f_{i,t} - \hat{f}_{\pi(i),t}|$, the deformation error as $\frac{1}{IT} \sum_{i,t} \|\varphi_t(x_i) - \hat{\varphi}_t(x_i)\|$, and the momenta error as $\frac{1}{IT} \sum_{i,t} \|p_{i,t} - \hat{p}_{\pi(i),t}\|$. The per-voxel log-likelihood is computed as $\frac{1}{|\Omega|T} \sum_{x,t} \log p(N_t(x) \mid \hat{\lambda}_t(x))$. Reported values correspond to the median, with 25th and 75th percentiles, over 100 independent MCMC runs.

Despite the heavy-tailed exponential prior with $\kappa=200$, the model achieves low fluorescence and template error in sparse regimes, indicating robustness to variability in signal strength. Template error decreases as the number of time points increases, suggesting that temporal coupling provides additional constraints that improve spatial localization and reduce ambiguity across frames. Recovery becomes more challenging as synapse density increases, since overlapping point spread functions introduce ambiguity in assigning fluorescence to nearby sources. This effect is illustrated in appendix Figure~\ref{fig:degenerate}, where closely spaced synapses lead to degraded recovery, reflecting a fundamental identifiability limitation of the imaging model.

\begin{figure}[!ht]
    \centering
    \includegraphics[width=0.95\linewidth]{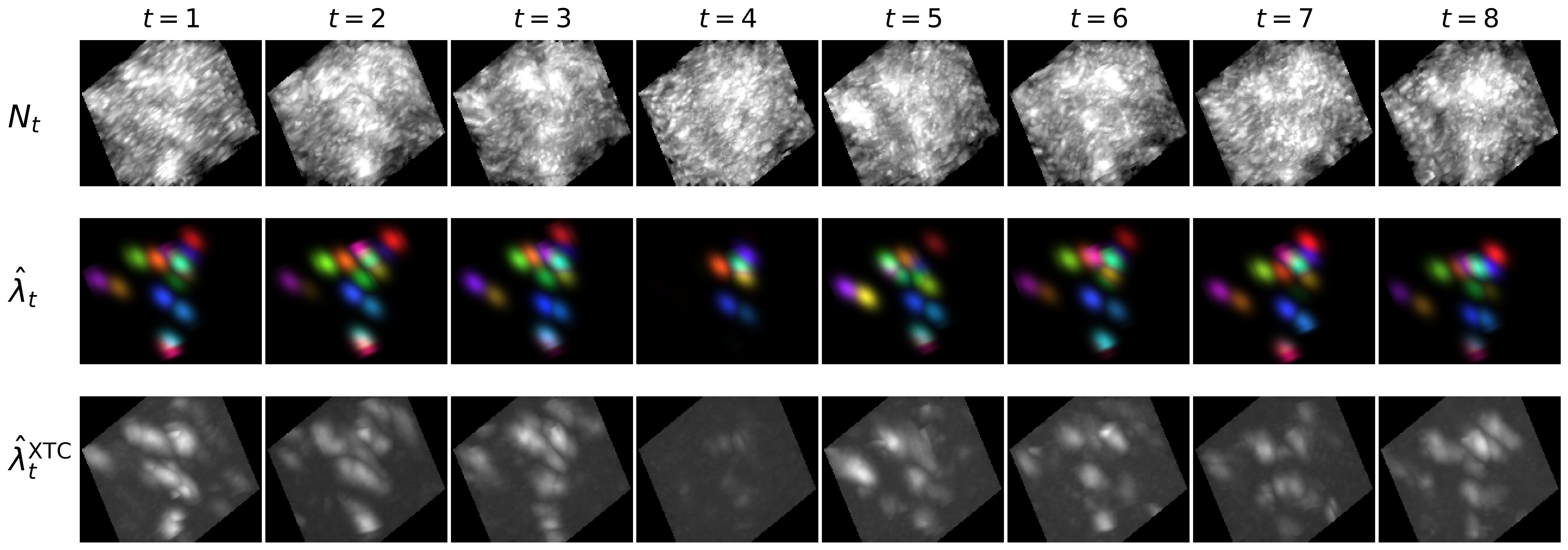}

\caption{Joint modeling produces temporally coherent intensity estimates.
Observed photon counts $N_t$ are noisy and vary across time.
Our method infers intensity $\hat{\lambda}_t$, yielding consistent synapse identities via a shared latent template and diffeomorphic transport.
The denoising autoencoder (XTC) produces visually similar intensity estimates $\hat{\lambda}^{\mathrm{XTC}}_t$, but lacks temporal consistency across frames.}
    \label{fig:sneak_peek_results}
\end{figure}

\subsection{Preliminary Results on tracking synapses in SEP-GluA2 transgenic mouse}

Figures \ref{fig:sneak_peek_results} show results on real data of synapses from~\citep{kumar2025uncertainty}. For all experiments, we denote our estimate of the latent foreground Poisson intensity by $\hat{\lambda}_t(x)$. We similarly interpret the XTC output as an estimate of foreground synaptic intensity, denoted $\hat{\lambda}_t^{\mathrm{XTC}}(x)$. There is strong visual agreement between our reconstruction $\hat{\lambda}_t(x)$ in Figure \ref{fig:sneak_peek_results}, row 2, and the denoising autoencoder baseline $\hat{\lambda}_t^{\mathrm{XTC}}(x)$~\citep{xu2023cross} in row 3. While both approaches recover similar spatial structure, a key difference is that our model produces a consistent set of synapse identities across time. Because the latent template is shared across all time points and transported via diffeomorphic transformations, tracking is handled implicitly within the model rather than requiring a separate detection and association step. This leads to trajectories that remain coherent across days, even in regions with closely spaced synapses where detection-based approaches are prone to merge or split errors.

\begin{figure*}[!ht]
    \centering
    \includegraphics[width=0.95\linewidth]{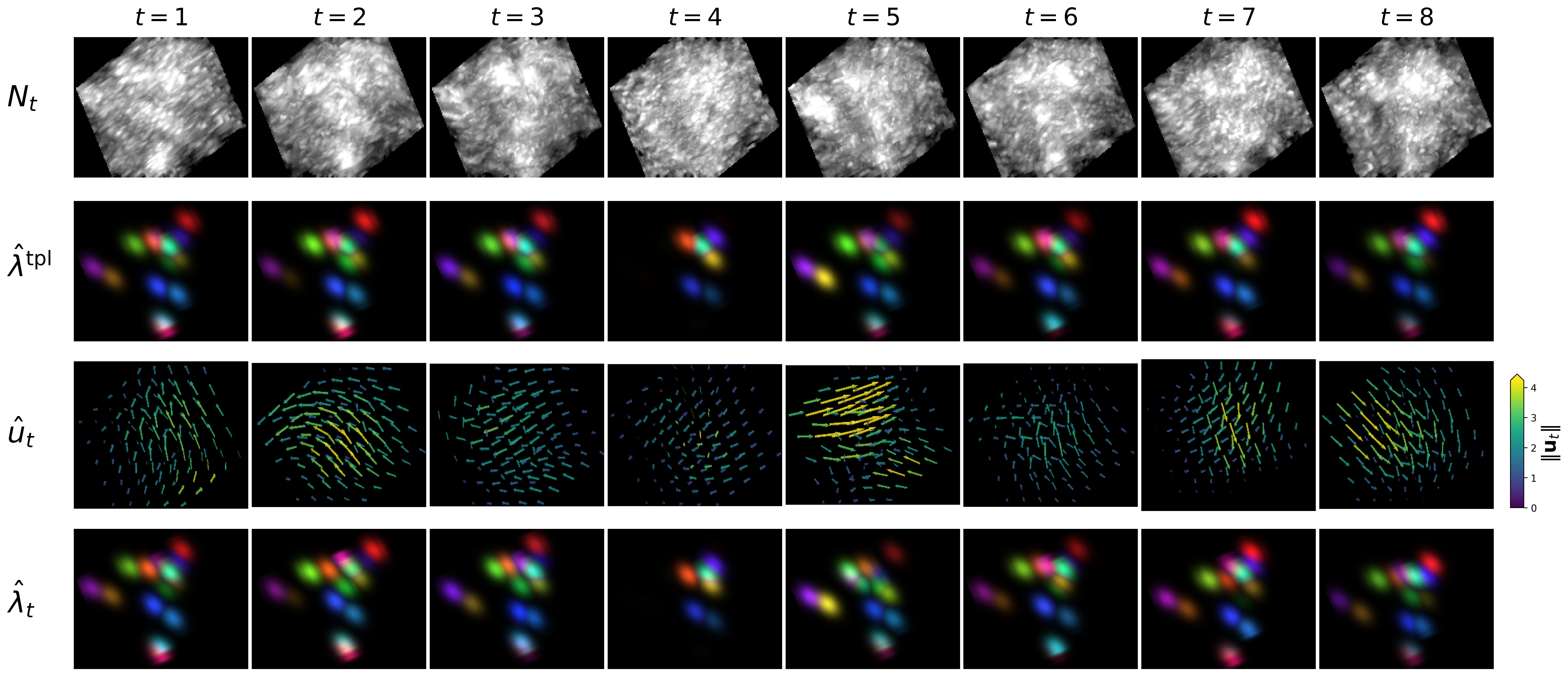}
    \caption{Stabilized Tracks, Displacement Fields and motion under the Displacement Field.}
    \label{fig:stabilization}
\end{figure*}

The inferred motion fields further reflect this consistency. The displacement fields $u_t(x)$ in Figure \ref{fig:stabilization} are smooth and do not exhibit self-intersections, which is a consequence of the diffeomorphic constraint. This prevents biologically implausible motion such as folding and ensures that synapse trajectories evolve smoothly over time. In practice, this results in more stable tracking and reduces identity switching across frames.

The posterior distributions in Figure \ref{fig:posterior_real} show tighter and more structured bands compared to the simulated setting. One possible explanation is that synapses in the 3D volume are less densely packed than in 2D projections, making it easier to resolve nearby sources and reducing ambiguity in the posterior.

In Table \ref{tab:real_data_comparison}, we provide a quantitative comparison with several baseline methods, including XTC~\citep{xu2023cross}, non-local means (NLM)~\citep{buades2005non}, and BM3D~\citep{dabov2007image}. Our model explicitly decomposes the reconstructed image into foreground synaptic intensity $\hat{\lambda}_t(x)$ and background fluorescence $\hat{\lambda}_0$. In contrast, baseline methods produce a single reconstructed intensity estimate without explicitly separating foreground and background components.

We first evaluate the mean voxel Poisson log-likelihood under the observation model $N_t(x) \sim \mathrm{Poisson}(\hat{\lambda}_t(x)+\hat{\lambda}_0)$, which measures how well the reconstructed intensity explains the observed photon counts under the imaging model. For baseline methods, their reconstructed output is treated as an estimate of the total image intensity. We next compute the root mean squared reconstruction error $\frac{1}{|\Omega|^{1/2}} \left\| N_t(x)-\hat{\lambda}_t(x)-\hat{\lambda}_0 \right\|_2$. To quantify temporal coherence, we compute the Pearson correlation between reconstructed foreground intensities from Day 1 and subsequent imaging sessions, $\mathrm{corr}(\hat{\lambda}_1(x), \hat{\lambda}_t(x))$ for $t=2,\dots,T$, averaged across all time points. Since baseline methods do not explicitly separate foreground and background fluorescence, their reconstructed outputs are used directly for this evaluation. Higher correlation indicates greater temporal stability of reconstructed synaptic structure after accounting for nonlinear tissue motion. Finally, we measure variability in the number of detected local peaks using a standard peak detection algorithm from Scipy~\citep{virtanen2020scipy} as a proxy for frame-to-frame detection stability.

\begin{table}[t]
\centering
\caption{
Quantitative comparison on real in vivo synapse data.
Our Bayesian model achieves higher Poisson likelihood, lower RMSE, and improved temporal consistency than XTC and is competitive with other baselines.
}
\small
\begin{tabular}{lcccc}
\toprule
\textbf{Metric} & \textbf{Bayesian (Ours)} & \textbf{XTC} & \textbf{NLM} & \textbf{BM3D} \\
\midrule
Poisson log-likelihood $\uparrow$ & -3.62 & -16.13 & -2.94 & -3.17 \\
RMSE $\downarrow$ & 8.33 & 10.76 & 3.62 & 5.73 \\
Temporal consistency $\uparrow$ & \textbf{0.577} & 0.387 & 0.316 & 0.454 \\
Peak stability (std) $\downarrow$ & \textbf{0.97} & 1.92 & 5.98 & 4.15 \\
\bottomrule
\end{tabular}
\label{tab:real_data_comparison}
\end{table}

The proposed method outperforms XTC in both Poisson likelihood and temporal consistency, indicating that explicitly modeling photon statistics and nonlinear tissue deformation improves both reconstruction fidelity and longitudinal stability. While NLM and BM3d achieve competitive RMSE through aggressive local smoothing and noise suppression, but they do not account for temporal correspondence or the underlying image formation process. In contrast, our framework jointly estimates signal, motion, and synapse identity directly from raw photon counts, producing temporally coherent reconstructions while remaining competitive in reconstruction error. Together, these results suggest that incorporating imaging physics and temporal structure provides a stronger inductive bias for longitudinal \textit{in vivo} synapse tracking under low-SNR and nonlinear deformation conditions.

\section{Discussion}

We presented an algorithm for estimating a template of point sources of fluorescence signal and tracking them over time directly from raw volumetric imaging data. By modeling synapses as latent point sources transported under diffeomorphic transformations, tracking emerges naturally from the generative model rather than requiring separate detection and data association steps. We introduced generalized momenta (latent cotangent variables) to parameterize diffeomorphic deformations of the domain, enabling flexible nonlinear motion while preserving topology. This allows the model to account for tissue deformation across time in a principled way and contributes to maintaining consistent synapse identities. The algorithm provides uncertainty estimates for all parameters, including synapse locations, fluorescence, and motion, while also producing direct estimates of fluorescence based on the physics of the microscope through a Poisson likelihood and PSF convolution. This joint modeling of signal, motion, and noise leads to more consistent and physically grounded reconstructions compared to pipeline-based approaches.

A key limitation of the current approach is computational cost, as MCMC inference can be slow for large volumes. In addition, recovery becomes challenging in regimes with densely packed synapses due to overlap in the point spread function, reflecting inherent ambiguity in the imaging process. Next, we aim to come up with better approximate the point spread function of the microscope, and to explore more efficient inference methods to scale this algorithm to larger brain volumes and longer time series.

\bibliographystyle{plainnat}
\bibliography{references}

@article{malinow2002ampa,
  title={AMPA receptor trafficking and synaptic plasticity},
  author={Malinow, Roberto and Malenka, Robert C},
  journal={Annual review of neuroscience},
  volume={25},
  number={1},
  pages={103--126},
  year={2002},
  publisher={Annual Reviews 4139 El Camino Way, PO Box 10139, Palo Alto, CA 94303-0139, USA}
}

@article{morris1989synaptic,
  title={Synaptic plasticity and learning II: do different kinds of plasticity underlie different kinds of learning?},
  author={Morris, Richard GM and Halliwell, Robert F and Bowery, N},
  journal={Neuropsychologia},
  volume={27},
  number={1},
  pages={41--59},
  year={1989},
  publisher={Elsevier}
}

@article{graves2021visualizing,
  title={Visualizing synaptic plasticity in vivo by large-scale imaging of endogenous AMPA receptors},
  author={Graves, Austin R and Roth, Richard H and Tan, Han L and Zhu, Qianwen and Bygrave, Alexei M and Lopez-Ortega, Elena and Hong, Ingie and Spiegel, Alina C and Johnson, Richard C and Vogelstein, Joshua T and  Tward, Daniel J and  Miller, Michael I and Huganir, Richard L},
  journal={Elife},
  volume={10},
  pages={e66809},
  year={2021},
  publisher={eLife Sciences Publications Limited}}

@article{chen2025automatic,
  author       = {Zhining Chen and Gabrielle I. Coste and Evan Li and Richard L. Huganir and Austin R. Graves and Adam S. Charles},
  title        = {Automatic detection of fluorescently labeled synapses in volumetric in vivo imaging data},
  journal      = {bioRxiv},
  year         = {2025},
  month        = {jan},
  day          = {23},
  doi          = {10.1101/2025.01.22.634278},
  note         = {Preprint},
  url          = {https://www.biorxiv.org/content/10.1101/2025.01.22.634278v1}
}

@article{brewer2013probabilistic,
  title={Probabilistic catalogs for crowded stellar fields},
  author={Brewer, Brendon J and Foreman-Mackey, Daniel and Hogg, David W},
  journal={The Astronomical Journal},
  volume={146},
  number={1},
  pages={7},
  year={2013},
  publisher={The American Astronomical Society}
}

@article{regier2019approximate,
  title={Approximate Inference for Constructing Astronomical Catalogs from Images},
  author={Regier, Jeffrey and Pamnany, Kiran and Giordano, Ryan and Thomas, Robert E and Schlegel, David and others},
  journal={The Annals of Applied Statistics},
  volume={13},
  number={4},
  pages={2450--2475},
  year={2019}
}

@article{abril2023pymc,
  title={PyMC: a modern, and comprehensive probabilistic programming framework in Python},
  author={Abril-Pla, Oriol and Andreani, Virgile and Carroll, Colin and Dong, Larry and Fonnesbeck, Christopher J and Kochurov, Maxim and Kumar, Ravin and Lao, Junpeng and Luhmann, Christian C and Martin, Osvaldo A and others},
  journal={PeerJ Computer Science},
  volume={9},
  pages={e1516},
  year={2023},
  publisher={PeerJ Inc.}
}

@article{grenander1998computational,
  title={Computational anatomy: An emerging discipline},
  author={Grenander, Ulf and Miller, Michael I},
  journal={Quarterly of applied mathematics},
  volume={56},
  number={4},
  pages={617--694},
  year={1998}
}

@article{kumar2025uncertainty,
  title={Uncertainty-Gated Min-Cost Flows for In Vivo NanoScale Synaptic Plasticity Tracking},
  author={Kumar, Shashwat and Coste, Gabrielle I and Premathilaka, Dasun and Huganir, Richard L and Graves, Austin R and Charles, Adam S and Miller, Michael I},
  journal={bioRxiv},
  pages={2025--10},
  year={2025},
  publisher={Cold Spring Harbor Laboratory}
}

@article{xu2023cross,
  title={Cross-modality supervised image restoration enables large-scale tracking of synaptic plasticity in vivo},
  author={Xu, Y. K. T. and others},
  journal={Nature},
  year={2023}
}

@article{denoyelle2017support,
  title={Support recovery for sparse super-resolution of positive measures},
  author={Denoyelle, Quentin and Duval, Vincent and Peyr{\'e}, Gabriel},
  journal={Journal of Fourier Analysis and Applications},
  volume={23},
  number={5},
  pages={1153--1194},
  year={2017}
}

@article{park2013variational,
  title={Variational semi-blind sparse deconvolution with orthogonal kernel bases and its application to MRFM},
  author={Park, Se Un and Dobigeon, Nicolas and Hero, Alfred O.},
  journal={IEEE Transactions on Image Processing},
  year={2013}
}

@article{preibisch2013efficient,
  title={Efficient Bayesian-based multiview deconvolution},
  author={Preibisch, Stephan and others},
  journal={Nature Methods},
  volume={11},
  number={6},
  pages={645--648},
  year={2014}
}

@article{ajdenbaum2023variational,
  title={A novel variational approach for multiphoton microscopy image restoration: from PSF estimation to 3D deconvolution},
  author={Ajdenbaum, Julien and others},
  journal={arXiv preprint arXiv:2311.18386},
  year={2023}
}

@article{trouve2005metamorphoses,
  title={Metamorphoses through lie group action},
  author={Trouv{\'e}, Alain and Younes, Laurent},
  journal={Foundations of computational mathematics},
  volume={5},
  number={2},
  pages={173--198},
  year={2005},
  publisher={Springer}
}

@article{roth2020cortical,
  title={Cortical synaptic AMPA receptor plasticity during motor learning},
  author={Roth, Richard H and Cudmore, Robert H and Tan, Han L and Hong, Ingie and Zhang, Yong and Huganir, Richard L},
  journal={Neuron},
  volume={105},
  number={5},
  pages={895--908},
  year={2020},
  publisher={Elsevier}
}

@article{winnubst2019reconstruction,
  title={Reconstruction of 1,000 projection neurons reveals new cell types and organization of long-range connectivity in the mouse brain},
  author={Winnubst, Johan and Bas, Erhan and Ferreira, Tiago A and Wu, Zhuhao and Economo, Michael N and Edson, Patrick and Arthur, Ben J and Bruns, Christopher and Rokicki, Konrad and Schauder, David and others},
  journal={Cell},
  volume={179},
  number={1},
  pages={268--281},
  year={2019},
  publisher={Elsevier}
}

@article{joshi2000landmark,
  title={Landmark matching via large deformation diffeomorphisms},
  author={Joshi, Sarang C and Miller, Michael I},
  journal={IEEE transactions on image processing},
  volume={9},
  number={8},
  pages={1357--1370},
  year={2000},
  publisher={IEEE}
}

@article{ebin1970groups,
  title={Groups of diffeomorphisms and the motion of an incompressible fluid},
  author={Ebin, David G. and Marsden, Jerrold},
  journal={Annals of Mathematics},
  volume={92},
  number={1},
  pages={102--163},
  year={1970},
  doi={10.2307/1970699}
}

@book{younes2010shapes,
  title={Shapes and Diffeomorphisms},
  author={Younes, Laurent},
  publisher={Springer},
  year={2010},
  doi={10.1007/978-3-642-12055-8}
}

@article{beg2005computing,
  title={Computing large deformation metric mappings via geodesic flows of diffeomorphisms},
  author={Beg, M. Faisal and Miller, Michael I. and Trouv{\'e}, Alain and Younes, Laurent},
  journal={International Journal of Computer Vision},
  volume={61},
  number={2},
  pages={139--157},
  year={2005},
  doi={10.1023/B:VISI.0000043755.93987.aa}
}

@inproceedings{glaunes2004diffeomorphic,
  title={Diffeomorphic matching of distributions: A new approach for unlabelled point-sets and sub-manifolds matching},
  author={Glaun{\`e}s, Joan and Trouv{\'e}, Alain and Younes, Laurent},
  booktitle={Proceedings of the 2004 IEEE Computer Society Conference on Computer Vision and Pattern Recognition},
  volume={2},
  pages={712--718},
  year={2004}
}

@article{buades2005non,
  title={A non-local algorithm for image denoising},
  author={Buades, Antoni and Coll, Bartomeu and Morel, Jean-Michel},
  journal={Proceedings of the IEEE Computer Society Conference on Computer Vision and Pattern Recognition (CVPR)},
  volume={2},
  pages={60--65},
  year={2005}
}

@article{dabov2007image,
  title={Image denoising by sparse 3-D transform-domain collaborative filtering},
  author={Dabov, Kostadin and Foi, Alessandro and Katkovnik, Vladimir and Egiazarian, Karen},
  journal={IEEE Transactions on Image Processing},
  volume={16},
  number={8},
  pages={2080--2095},
  year={2007}
}

@article{virtanen2020scipy,
  title={SciPy 1.0: Fundamental Algorithms for Scientific Computing in Python},
  author={Virtanen, Pauli and Gommers, Ralf and Oliphant, Travis E. and Haberland, Matt and Reddy, Tyler and Cournapeau, David and Burovski, Evgeni and Peterson, Pearu and Weckesser, Warren and Bright, Jonathan and others},
  journal={Nature Methods},
  volume={17},
  number={3},
  pages={261--272},
  year={2020}
}



\newpage

\clearpage
\newpage
\appendix
\section{Appendix}
\begin{figure*}[!ht]\includegraphics[
  width=\linewidth]{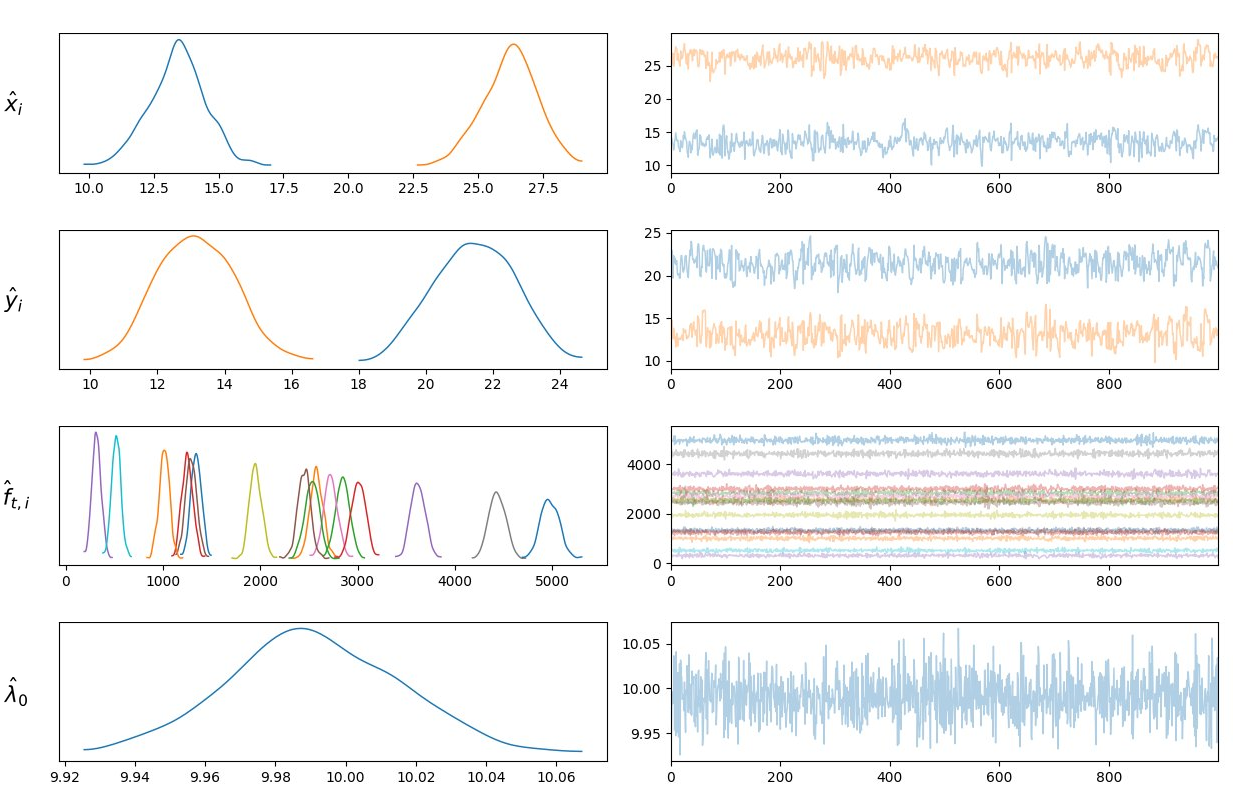}
\caption{Posterior distributions concentrate around ground truth.
Marginal posteriors over template locations, fluorescence parameters, motion variables, and background intensity exhibit well-defined modes and limited spread, indicating stable inference and good identifiability in the sparse regime.}
    \label{fig:simulate_posterior}
\end{figure*}

\begin{figure*}[!ht]
    \centering

    \includegraphics[width=0.95\linewidth]{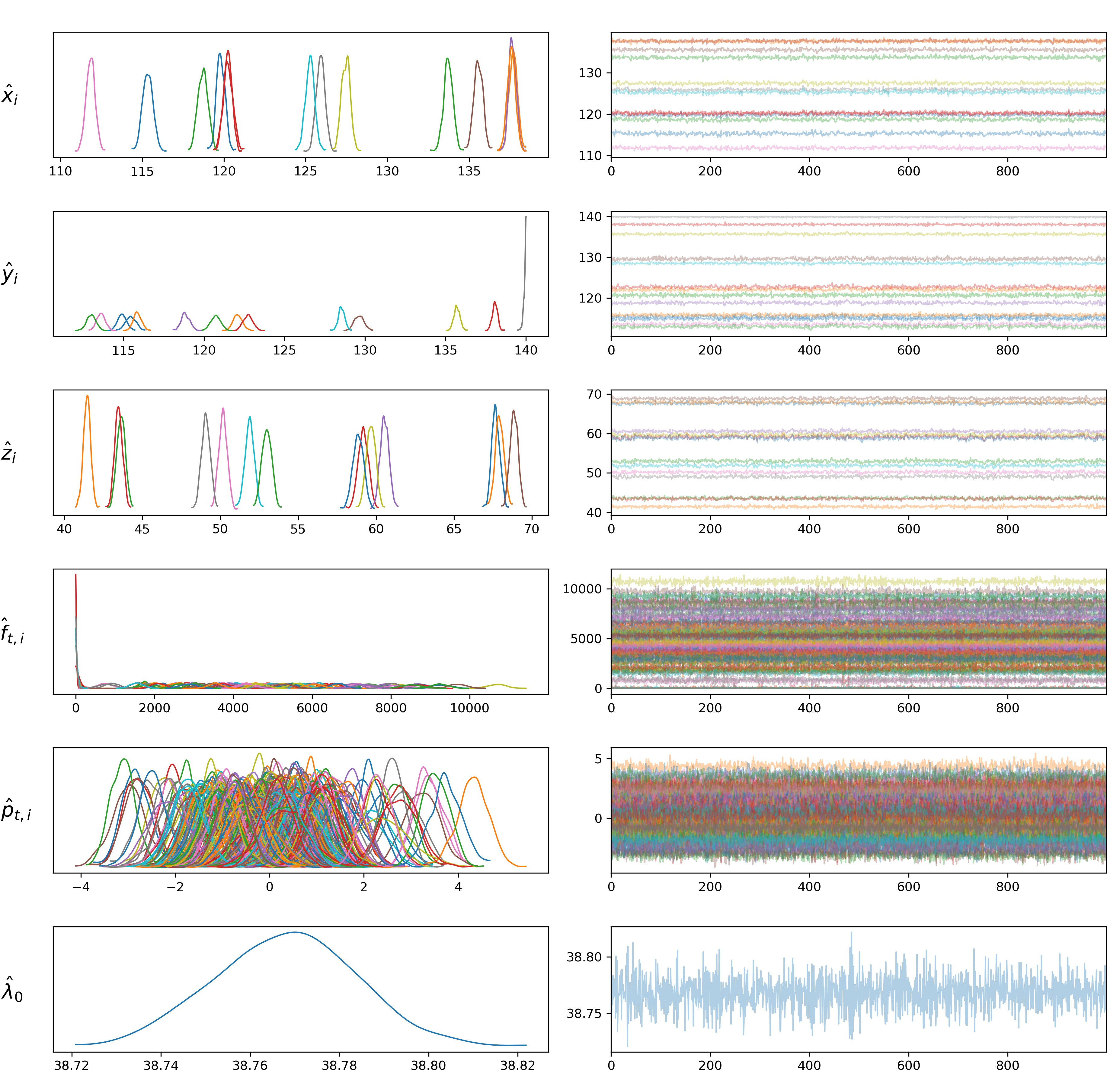}
    \caption{Posterior distributions obtained from model.}
    \label{fig:posterior_real}
\end{figure*}

\subsection{Gradient derivations}
We have the posterior
\begin{equation}
    \log \pi(\theta|N) = \sum_{t=1}^T \sum_{x \in \Omega}  (N_t(x) \log (\lambda_t(x) + \lambda_0) - (\lambda_t(x) + \lambda_0))+ \log \pi_0(\theta),
\end{equation}
where $\pi_0$ denotes the chosen prior models on components of $\theta$. 
Taking a partial derivative w.r.t a parameter $\theta$ by the chain rule:

\begin{align}
    \frac{\partial \log \pi(\theta|N)}{\partial \theta} &= \frac{\partial \log \pi(\theta|N)}{\partial \lambda_t(x)} \frac{\partial \lambda_t(x)}{\partial \theta} \nonumber \\
    &= \left( \sum_{t=1}^T \sum_{x \in \Omega} r_t(x, \lambda_0) \frac{\partial \lambda_t(x)}{\partial \theta} + \frac{\partial \log \pi_0}{\partial \theta} \right) \ \ \in \real^m,
    \label{eq:gradient_chain_rule}
\end{align}
with   $r_t(x, \lambda_0)=\frac{N_t(x)}{\lambda_t(x) + \lambda_0} - 1\ \ \in \real$.

\subsubsection{Derivative of $U$ with respect to $f_{i,t'}$}
Since $\lambda_t(x) =  \sum_{i'=1}^I f_{i',t} \psi(x-\varphi_t(x_i'))$, its derivative is:  
\begin{align}
    \frac{\partial \lambda_t(x)}{\partial f_{i,t'}} &= \frac{\partial}{\partial f_{i,t'}}(\sum_{i'=1}^I f_{i',t}\psi(x-\varphi_t(x_{i'})))
    = \psi(x-\varphi_t(x_i)) \delta(t,t')\ \in \real, \\
    \frac{\partial \log \pi_0}{\partial f_{i,t'}} &= \frac{\partial}{\partial f_{i,t'}}(\log \kappa - \kappa f_{i,t'}) = -\kappa \ \ \in \real.
\end{align}

Substituting in \eqref{eq:gradient_chain_rule} and simplifying the Kronecker delta, we get
\begin{align}
    \frac{\partial \log \pi(\theta|N)}{\partial f_{i,t'}} &= \left( \sum_{x \in \Omega} r_{t'}(x, \lambda_0) \psi(x-\varphi_{t'}(x_i)) - \kappa \right) \ \ \in \real. 
\end{align}

\subsubsection{Derivative of $U$ with respect to $p_{j,t'}$}

\begin{align}
    \frac{\partial \lambda_t(x)}{\partial p_{j,t'}} &= \frac{\partial}{\partial p_{j,t'}}(\sum_{i=1}^I f_{i,t}\psi(x-\varphi_t(x_{i}))) \nonumber \\
    &= \sum_{i=1}^I f_{i,t} \nabla_x \psi(x-\varphi_t(x_i))^T \frac{\partial (x - \varphi_t(x_i))}{\partial p_{j,t'}} \nonumber \\
    &= \sum_{i=1}^I f_{i,t} \nabla_x \psi(x-\varphi_t(x_i))^T \frac{-\partial \varphi_t(x_i)}{\partial p_{j,t'}} \ \ \in \real^3\ .
\end{align}
Now,
\begin{align}
    \frac{\partial \varphi_t(x_i)}{\partial p_{j,t'}} &= \frac{\partial }{\partial p_{j,t'}} (x_i + \sum_{i'=1}^I k(x_i,x_{i'}) p_{i',t}) \nonumber \\
    &= I_3 k(x_i,x_j)  \delta(t',t) \ \ \in \real^{3 \times 3}.
\end{align}

\begin{align}
    \frac{\partial \log \pi_0}{\partial p_{j,t'}} = \frac{\partial}{\partial p_{j,t'}}(-\frac{1}{2} p_{j,t'}^T \Sigma_{\text{motion}}^{-1} p_{j,t'}) = -p_{j,t'}^T\Sigma_{\text{motion}}^{-1} \ \ \in \real^{3}
\end{align}

Substituting in \eqref{eq:gradient_chain_rule} and simplifying the Kronecker delta, we get
\begin{align}
    \frac{\partial \log \pi(\theta|N)}{\partial p_{j,t'}} &= \sum_{x \in \Omega} r_{t'}(x) \sum_{i=1}^I f_{i,t'} \nabla \psi(x-\varphi_{t'}(x_i))^T \frac{-\partial \varphi_{t'}(x_i)}{\partial p_{j,t'}} - p_{j,t'}^T \Sigma_{\text{motion}}^{-1} \nonumber \\
    &= -\sum_{x \in \Omega} r_{t'}(x) \sum_{i=1}^I f_{i,t'} \nabla_x \psi(x-\varphi_{t'}(x_i))^T k(x_i,x_j) -  p_{j,t'}^T\Sigma_{\text{motion}}^{-1} \ \ \in \real^3.
\end{align}

\subsubsection{Derivative of $U$ with respect to $x_i$}
\begin{align}
    \frac{\partial \lambda_t(x)}{\partial x_j} &=  \frac{\partial}{\partial x_j}(\sum_{i=1}^I f_{i,t}\psi(x-\varphi_t(x_{i})) \nonumber \\
    &=  f_{j,t} \frac{\partial}{\partial x_j} \psi(x-\varphi_t(x_{j})) 
    = -f_{j,t} \nabla \psi(x-\varphi_{t}(x_j))^T \frac{\partial \varphi_t(x_j)}{\partial x_j}
\end{align}
Using the definition of $\varphi_t(x)$, we get:
\begin{align}
    \frac{\partial \varphi_t(x_j)}{\partial x_j} &= \frac{\partial }{\partial x_j} (x_j + \sum_{i'=1}^I k(x_j,x_{i'}) p_{i',t'})
    = I + \sum_{i'=1}^I \nabla_{1} k(x_j,x_{i'}) p_{i',t'}^T\ \ \in \real^{3 \times 3}.
\end{align}
Substituting in \eqref{eq:gradient_chain_rule}, we get
\begin{align}
    \frac{\partial \log \pi(\theta|N)}{\partial x_{j}} &= \sum_{t=1}^T\sum_{x \in \Omega} r_t(x, \lambda_0) f_{j,t} \nabla \psi(x-\varphi_{t}(x_j))^T \frac{-\partial \varphi_{t}(x_j)}{\partial x_j} \nonumber \\
    &= -\sum_{t=1}^T\sum_{x \in \Omega} r_t(x, \lambda_0) f_{j,t} \nabla_x \psi(x-\varphi_{t}(x_j))^T (I_3 + \sum_{i'=1}^I \nabla_1 k(x_j,x_{i'}) p_{i',t}^T) \ \ \in \real^3. 
\end{align}

\subsection{Derivative of $U$ with respect to $\lambda_0$}
\begin{align}
    \frac{\partial \log \pi(\theta|N)}{\partial \lambda_0} &=\sum_{t=1}^T \sum_{x \in \Omega} r_t(x) - \frac{\lambda_0}{\nu^2}
    \label{eq:gradient_lambda_0} \ \ \in \real.
\end{align}

\section{Compute resources.}
All experiments were performed on a local Ubuntu workstation with an AMD Ryzen 7 9800X3D 8-Core CPU (16 threads) and 30 GB RAM. Inference was implemented using PyMC 5.28.4 and executed entirely on CPU without GPU acceleration. Typical simulated experiments required approximately 15 minutes per run. Real-data MCMC inference on a small $30 \times 30 \times 30$ voxel subvolume required approximately one day of runtime, reflecting the computational cost of high-dimensional Bayesian posterior sampling with diffeomorphic motion and Poisson image formation.

\begin{figure*}[!ht]
\includegraphics[
  width=\linewidth]{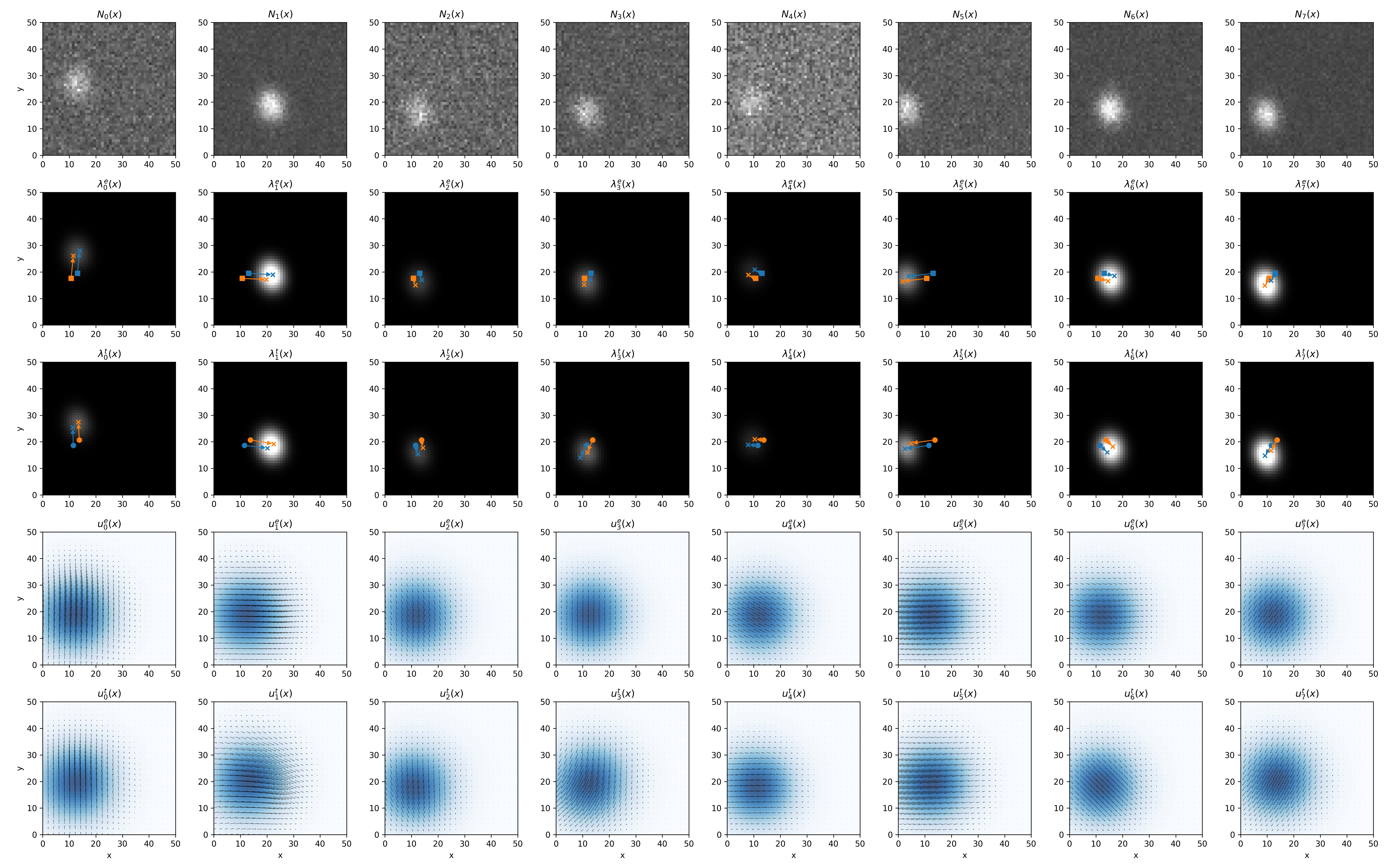}
  \includegraphics[
  width=\linewidth]{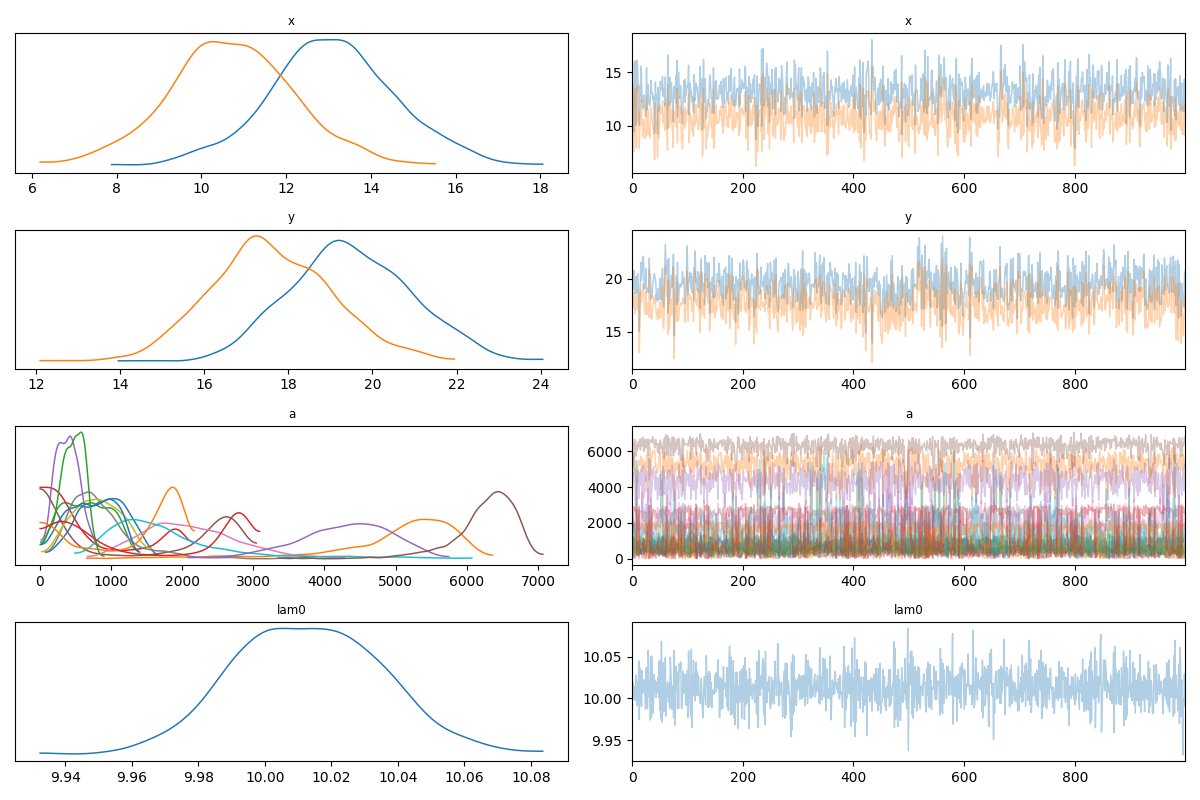}
    \caption{Degenerate case in Simulated data. Fluorescence error: 697/2000.}
    \label{fig:degenerate}
\end{figure*}


\clearpage
\section*{NeurIPS Paper Checklist}

\begin{enumerate}

\item {\bf Claims}
    \item[] Question: Do the main claims made in the abstract and introduction accurately reflect the paper's contributions and scope?
    \item[] Answer: \answerYes{}
    \item[] Justification: The abstract and introduction clearly state the main contributions: a joint Bayesian framework for synapse localization and tracking, integration of diffeomorphic registration into inference, uncertainty quantification, and validation on simulated and real longitudinal microscopy datasets.
    \item[] Guidelines:
    \begin{itemize}
        \item The answer \answerNA{} means that the abstract and introduction do not include the claims made in the paper.
        \item The abstract and/or introduction should clearly state the claims made, including the contributions made in the paper and important assumptions and limitations. A \answerNo{} or \answerNA{} answer to this question will not be perceived well by the reviewers.
        \item The claims made should match theoretical and experimental results, and reflect how much the results can be expected to generalize to other settings.
        \item It is fine to include aspirational goals as motivation as long as it is clear that these goals are not attained by the paper.
    \end{itemize}

\item {\bf Limitations}
    \item[] Question: Does the paper discuss the limitations of the work performed by the authors?
    \item[] Answer: \answerYes{}
    \item[] Justification: The Discussion section describes limitations including computational cost of MCMC inference, scalability challenges for large volumes, and ambiguity caused by densely packed synapses and overlapping point spread functions.
    \item[] Guidelines:
    \begin{itemize}
        \item The answer \answerNA{} means that the paper has no limitation while the answer \answerNo{} means that the paper has limitations, but those are not discussed in the paper.
        \item The authors are encouraged to create a separate ``Limitations'' section in their paper.
        \item The paper should point out any strong assumptions and how robust the results are to violations of these assumptions.
        \item The authors should reflect on the scope of the claims made.
        \item The authors should discuss the computational efficiency of the proposed algorithms and how they scale with dataset size.
        \item Reviewers will be specifically instructed to not penalize honesty concerning limitations.
    \end{itemize}

\item {\bf Theory assumptions and proofs}
    \item[] Question: For each theoretical result, does the paper provide the full set of assumptions and a complete (and correct) proof?
    \item[] Answer: \answerYes{}
    \item[] Justification: The paper explicitly states the probabilistic assumptions, priors, likelihood formulation, and derivations for posterior and HMC gradients.
    \item[] Guidelines:
    \begin{itemize}
        \item The answer \answerNA{} means that the paper does not include theoretical results.
        \item All theorems, formulas, and proofs in the paper should be numbered and cross-referenced.
        \item All assumptions should be clearly stated or referenced.
        \item Informal proofs should be complemented by formal derivations where appropriate.
    \end{itemize}

\item {\bf Experimental result reproducibility}
    \item[] Question: Does the paper fully disclose all the information needed to reproduce the main experimental results?
    \item[] Answer: \answerYes{}
    \item[] Justification: Experimental settings, priors, covariance parameters, evaluation metrics, inference settings, and dataset descriptions are provided for both simulated and real experiments.
    \item[] Guidelines:
    \begin{itemize}
        \item The answer \answerNA{} means that the paper does not include experiments.
        \item Making the paper reproducible is important regardless of whether the code and data are provided.
        \item Reproducibility may be supported through detailed instructions, model access, or released code/data.
    \end{itemize}

\item {\bf Open access to data and code}
    \item[] Question: Does the paper provide open access to the data and code?
    \item[] Answer: \answerNo{}
    \item[] Justification: Code is not included in the current submission. We plan to release the implementation upon acceptance.
    \item[] Guidelines:
    \begin{itemize}
        \item While release of code and data is encouraged, \answerNo{} is acceptable if justified.
        \item Instructions should ideally contain commands and environments needed to reproduce the experiments.
    \end{itemize}

\item {\bf Experimental setting/details}
    \item[] Question: Does the paper specify all the training and test details necessary to understand the results?
    \item[] Answer: \answerYes{}
    \item[] Justification: The manuscript specifies simulation parameters, HMC settings, imaging setup, evaluation metrics, and reconstruction procedures necessary for reproducing the experiments.
    \item[] Guidelines:
    \begin{itemize}
        \item The experimental setting should be presented in enough detail to understand the reported results.
        \item Full details may appear in the appendix or supplemental material.
    \end{itemize}

\item {\bf Experiment statistical significance}
    \item[] Question: Does the paper report error bars suitably and correctly defined or other appropriate information about statistical significance?
    \item[] Answer: \answerYes{}
    \item[] Justification: Simulated experiments report medians together with 25th and 75th percentiles. Posterior distributions obtained from MCMC are also visualized.
    \item[] Guidelines:
    \begin{itemize}
        \item Error bars, confidence intervals, or posterior uncertainty estimates should be clearly defined.
        \item The factors of variability captured by the uncertainty estimates should be stated.
    \end{itemize}

\item {\bf Experiments compute resources}
    \item[] Question: For each experiment, does the paper provide sufficient information on the computer resources needed to reproduce the experiments?
    \item[] Answer: \answerYes{}
    \item[] Justification: Hardware details are provided in the section on Compute resources in appendix.
    \item[] Guidelines:
    \begin{itemize}
        \item The paper should ideally disclose compute hardware, memory requirements, and approximate execution time.
    \end{itemize}

\item {\bf Code of ethics}
    \item[] Question: Does the research conducted in the paper conform, in every respect, with the NeurIPS Code of Ethics?
    \item[] Answer: \answerYes{}
    \item[] Justification: This work focuses on computational analysis of neuroscience imaging data and does not present immediate societal or ethical risks.
    \item[] Guidelines:
    \begin{itemize}
        \item Authors should ensure compliance with the NeurIPS Code of Ethics and preserve anonymity where required.
    \end{itemize}

\item {\bf Broader impacts}
    \item[] Question: Does the paper discuss both potential positive societal impacts and negative societal impacts of the work performed?
    \item[] Answer: \answerYes{}
    \item[] Justification: The paper discusses the potential positive impact of improved quantitative analysis of synaptic plasticity and longitudinal neuroscience imaging for biomedical research applications, including studies of learning, memory, and neurological disease. Potential negative impacts are limited because the work is intended for scientific imaging applications rather than deployment in consumer-facing systems. The paper also notes that inaccurate probabilistic reconstructions or modeling assumptions could bias downstream biological interpretation if uncertainty estimates are ignored or insufficiently validated.
    \item[] Guidelines:
    \begin{itemize}
        \item Authors should discuss both beneficial applications and possible risks or unintended consequences where applicable.
    \end{itemize}

\item {\bf Safeguards}
    \item[] Question: Does the paper describe safeguards for responsible release of data or models with high misuse potential?
    \item[] Answer: \answerNA{}
    \item[] Justification: The paper does not release high-risk generative models, scraped datasets, or systems with substantial misuse potential.
    \item[] Guidelines:
    \begin{itemize}
        \item The answer \answerNA{} means that the paper poses no substantial misuse risk.
    \end{itemize}

\item {\bf Licenses for existing assets}
    \item[] Question: Are the creators or original owners of assets used in the paper properly credited?
    \item[] Answer: \answerYes{}
    \item[] Justification: Existing datasets, preprocessing pipelines, and prior methods used in this work are appropriately cited in the manuscript.
    \item[] Guidelines:
    \begin{itemize}
        \item Existing datasets, codebases, and models should be properly cited and attributed.
    \end{itemize}

\item {\bf New assets}
    \item[] Question: Are new assets introduced in the paper well documented?
    \item[] Answer: \answerNA{}
    \item[] Justification: The paper does not currently release new datasets, code, or pretrained models as part of the submission.
    \item[] Guidelines:
    \begin{itemize}
        \item Released assets should include documentation, limitations, and licensing information where applicable.
    \end{itemize}

\item {\bf Crowdsourcing and research with human subjects}
    \item[] Question: For crowdsourcing experiments and research with human subjects, does the paper include instructions and compensation details?
    \item[] Answer: \answerNA{}
    \item[] Justification: The paper does not involve crowdsourcing or research with human subjects.
    \item[] Guidelines:
    \begin{itemize}
        \item The answer \answerNA{} means that the paper does not involve crowdsourcing or human subject studies.
    \end{itemize}

\item {\bf Institutional review board (IRB) approvals or equivalent for research with human subjects}
    \item[] Question: Does the paper describe risks to participants and IRB approvals where applicable?
    \item[] Answer: \answerNA{}
    \item[] Justification: The paper does not involve human subjects research.
    \item[] Guidelines:
    \begin{itemize}
        \item The answer \answerNA{} means that the paper does not involve crowdsourcing or human subjects research.
    \end{itemize}

\item {\bf Declaration of LLM usage}
    \item[] Question: Does the paper describe usage of LLMs if they are an important or non-standard component of the research?
    \item[] Answer: \answerNA{}
    \item[] Justification: LLMs are not part of the core methodology, experiments, or scientific contributions of this work.
    \item[] Guidelines:
    \begin{itemize}
        \item The answer \answerNA{} means that LLMs are not part of the core methodology of the research.
    \end{itemize}

\end{enumerate}
\end{document}